\newcommand{\CLASSINPUTtoptextmargin}{2.4cm}
\newcommand{\CLASSINPUTbottomtextmargin}{4.6cm}
\documentclass[conference]{IEEEtran}
\IEEEoverridecommandlockouts
\usepackage{cite}
\usepackage{amsmath,amssymb,amsfonts}
\usepackage{algorithmic}
\usepackage[ruled,vlined]{algorithm2e}
\usepackage{graphicx}
\usepackage{textcomp}
\usepackage{xcolor}
\usepackage{tabularx}
\usepackage{booktabs}
\usepackage{graphicx}
\usepackage{subcaption}
\usepackage{multirow}
\usepackage{tabularray}
\usepackage{color}
\usepackage{pdfcomment}
\usepackage{todonotes}
\usepackage{placeins}

\def\BibTeX{{\rm B\kern-.05em{\sc i\kern-.025em b}\kern-.08em
    T\kern-.1667em\lower.7ex\hbox{E}\kern-.125emX}}

\begin{document}

\title{
Plug-and-Play AMC: Context Is King in Training-Free, Open-Set Modulation with LLMs\\
}

\author{\IEEEauthorblockN{1\textsuperscript{st} Mohammad Rostami}
\IEEEauthorblockA{\textit{Electrical and Computer Engineering} \\
\textit{Rowan University}\\
Glassboro, NJ  \\
rostami23@students.rowan.edu}
\and
\IEEEauthorblockN{2\textsuperscript{nd} Atik Faysal}
\IEEEauthorblockA{\textit{Electrical and Computer Engineering} \\
\textit{Rowan University}\\
Glassboro, NJ \\
faysal24@rowan.edu}
\and
\IEEEauthorblockN{3\textsuperscript{rd} Reihaneh Gh. Roshan}
\IEEEauthorblockA{\textit{Computer Science} \\
\textit{Stevens Institute of Technology}\\
Hoboken, NJ \\
rghasemi@stevens.edu}
\and
\IEEEauthorblockN{4\textsuperscript{th} Huaxia Wang}
\IEEEauthorblockA{\textit{Electrical and Computer Engineering} \\
\textit{Rowan University}\\
Glassboro, NJ  \\
wanghu@rowan.edu}
\and
\IEEEauthorblockN{5\textsuperscript{th} Nikhil Muralidhar}
\IEEEauthorblockA{\textit{Computer Science} \\
\textit{ Stevens Institute of Technology}\\
Hoboken, NJ \\
nmurali1@stevens.edu}
\and
\IEEEauthorblockN{6\textsuperscript{th} Yu-Dong Yao}
\IEEEauthorblockA{\textit{Electrical and Computer Engineering} \\
\textit{ Stevens Institute of Technology}\\
Hoboken, NJ \\
yyao@stevens.edu}
}

\newcommand{\nikhilc}[1]{\textcolor{red}{\,Nikhil says: #1}}
\newcommand\blfootnote[1]{%
  \begin{NoHyper}%
  \renewcommand\thefootnote{}\footnote{#1}%
  \addtocounter{footnote}{-1}%
  \end{NoHyper}%
}

\maketitle

\begin{abstract}
    Automatic Modulation Classification (AMC) is critical for efficient spectrum management and robust wireless communications. However, AMC remains challenging due to the complex interplay of signal interference and noise. In this work, we propose an innovative framework that integrates traditional signal processing techniques with Large-Language Models (LLMs) to address AMC. Our approach leverages higher-order statistics and cumulant estimation to convert quantitative signal features into structured natural language prompts. By incorporating exemplar contexts into these prompts, our method exploits the LLM's inherent familiarity with classical signal processing, enabling effective one-shot classification without additional training or preprocessing (e.g., denoising). Experimental evaluations on synthetically generated datasets—spanning both noiseless and noisy conditions—demonstrate that our framework achieves competitive performance across diverse modulation schemes and Signal-to-Noise Ratios (SNRs). Moreover, our approach paves the way for robust foundation models in wireless communications across varying channel conditions, significantly reducing the expense associated with developing channel-specific models. This work lays the foundation for scalable, interpretable, and versatile signal classification systems in next-generation wireless networks. The source code is available at \url{https://github.com/RU-SIT/context-is-king}
\end{abstract}


\section{Introduction}
\IEEEPARstart{A}{utomatic} Modulation Classification (AMC) is a critical technology in modern wireless communications, underpinning applications in cognitive radio, spectrum sensing, and interference management. By accurately identifying modulation schemes, AMC facilitates efficient spectrum utilization and enhances the adaptability and reliability of communication networks. However, despite its significance, AMC remains a challenging problem due to the complex interplay of signals with ambient noise and other channel impairments.

Early approaches to AMC predominantly relied on traditional machine learning techniques and handcrafted feature extraction, but these methods have gradually been superseded by deep learning models. For instance, Convolutional Neural Networks (CNNs) have been extensively applied in this domain. Peng et al.~\cite{peng2017modulation} transformed raw modulated signals into constellation diagrams to serve as inputs for CNN architectures such as AlexNet, while subsequent studies introduced variants like MCNet~\cite{8963964} that leveraged multiple convolutional blocks with skip connections and asymmetric kernels to capture spatio-temporal correlations. In another investigation, a CNN framework was explored using multiple signal representations, such as constellation diagrams, ambiguity functions, and eye diagrams, with empirical results indicating that combining these representations could improve classification accuracy~\cite{10139474}.

\begin{figure*}
    \centering
    \includegraphics[width=0.9\textwidth]{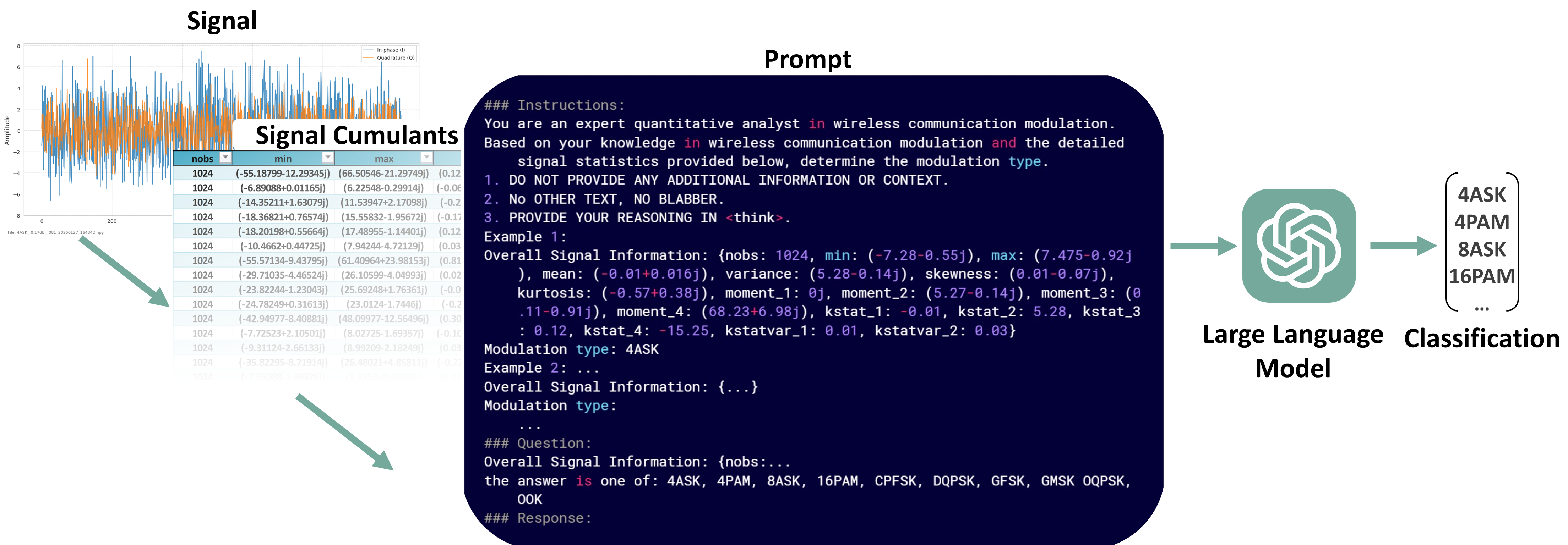}
    \caption{Overview of our in-context AMC framework. The raw signal is first summarized into high-order statistics and cumulants, which are then embedded in a text prompt. This prompt is fed into a large language model to perform modulation classification from a predefined set of modulation types (e.g., 4ASK, 4PAM, 8ASK, 16PAM).}
    \label{fig:method}
\end{figure*}

More recently, attention-based models, particularly transformers, have garnered significant interest for AMC. Transformers, introduced by Vaswani et al.~\cite{vaswani2017attention}, are adept at processing sequential data by dynamically focusing on the most relevant parts of the input. Cai et al.~\cite{cai2022signal} demonstrated that a transformer network (TRN)-based approach can capture long-range dependencies more effectively than conventional CNNs, which is especially beneficial in low SNR scenarios. Building on this idea, Kong et al.~\cite{kong2023transformer} further improved transformer-based AMC by incorporating convolutional embedding and attention pooling techniques, and Qu et al.~\cite{qu2024enhancing} combined transformers with Long Short-Term Memory (LSTM) networks to address long-range temporal dependencies and enhance robustness against noise through novel data augmentation strategies.

Despite these advancements, many existing methods are developed under idealized, noise-free conditions~\cite{peng2018modulation,8454504}. Such conditions are rarely encountered in real-world environments where signals are typically contaminated with noise. Moreover, several approaches struggle with fine-tuning downstream tasks when only limited labeled data is available~\cite{hong2017automatic,ramjee2019fast}.

Nmformer \cite{faysal2024nmformer} employs a Vision Transformer (ViT) trained on a large labeled dataset to achieve strong performance when retrained on low-SNR signals. However, the model struggles to attain high accuracy due to the absence of unsupervised pretraining. In contrast, DenoMAE \cite{faysal2025denomae} and DenoMAE2.0 \cite{faysal2025denomae20improvingdenoisingmasked} aim to achieve competitive performance with only a few thousand samples during pretraining and significantly fewer labeled samples in the downstream task. Nevertheless, the limited data constrains the model's ability to achieve substantial performance improvements. This underscores the need for a more advanced framework that effectively leverages large-scale data to enhance accuracy.

Addressing these limitations, recent work has begun to explore more sophisticated frameworks that can balance performance across varying SNR conditions. Gao et al. in the MoE-AMC study~\cite{gao2023moe} proposed a Mixture-of-Experts (MoE) approach that integrates a ResNet-based model for high SNR signals with a transformer-based model for low SNR signals, achieving a more balanced and robust classification performance. Complementing this, the review by Jafarigol et al.~\cite{jafarigol2025ai} provided a comprehensive analysis of contemporary AI/ML-based AMC models, highlighting both recent advances and persistent challenges, particularly the difficulty of maintaining accuracy in noisy environments and the need for efficient models that operate under constrained computational resources.

In this work, we take a novel direction by leveraging a pretrained LLM-based approach for automatic modulation classification. Our method utilizes traditional modulation statistics and cumulants, capitalizing on the LLM’s inherent familiarity with classical signal processing techniques. This capability enables our framework to perform competitive one-shot modulation classification without the need for exhaustive training across all SNRs, modulation types, or channel conditions, and without the computational overhead of extra preprocessing steps such as denoising. By transforming quantitative signal features into natural language descriptions, our approach bridges the gap between classical signal processing and modern AI, achieving robust performance even in challenging, noisy scenarios.

Our contributions include:
\begin{itemize}
    \item A novel in-context approach for AMC using LLMs that simplifies the processing pipeline.
    \item A one-shot classification method that eliminates the need for extensive training data.
    \item Competitive performance under both noiseless and noisy conditions.
\end{itemize}

Beyond improving AMC performance, our approach paves the way for developing foundation models for wireless communications. Such foundation models can provide a robust unified solution across diverse scenarios, reducing the need for channel-specific models and lowering the associated development and deployment costs. By enabling models that generalize well without extensive retraining, our method has the potential to streamline the design of wireless communication systems, making them more adaptable, scalable, and cost-effective.

Overall, this work not only simplifies modulation classification but also sets the stage for further research in robust, scalable, and versatile signal processing using large language models.


\section{Theoretical Background}
\label{sec:back}
This section presents the key theoretical foundations of higher-order statistics and cumulants that capture essential signal features and inform our LLM prompt construction approach.

\subsection{High-Order Statistics}
Statistical moments offer a quantitative description of the shape and characteristics of a random variable's probability distribution. In particular, the $n^{\mathrm{th}}$-order moment of a random variable $x$ is defined as
\begin{equation}
    m_n(x) = \int x^n f(x) \, dx,
\end{equation}
where $f(x)$ represents the probability density function of $x$. The existence of the $n^{\mathrm{th}}$-order moment guarantees that all moments of lower order are also defined. We refer to $m_n(x-\mu)$ as the \emph{central moment}, where $\mu$ is the mean, and $m_n(x/\sigma)$ as the \emph{normalized moment}, with $\sigma$ denoting the standard deviation.

Cumulants, denoted by $c_n$, provide an alternative distribution characterization. While the first three cumulants correspond to the mean, variance, and skewness, respectively, cumulants of order four and above capture additional aspects of the distribution's shape that are not directly reflected in the moments. The cumulant-generating function is given by
\begin{equation}
    C_t(x) = \log\left(E\left[e^{tx}\right]\right),
\end{equation}
which can be expanded into a power series as
\begin{equation}
    C_t(x) = \sum_{n=1}^{\infty} \frac{c_n t^n}{n!}.
\end{equation}

By differentiating $C_t(x)$ $n$ times and evaluating at $t=0$, one can obtain the $n^{\mathrm{th}}$ cumulant. A noteworthy property of cumulants is their additive behavior for independent random variables:
\begin{equation}
    c_n(X+Y) = c_n(X) + c_n(Y).
\end{equation}

In practical applications, cumulants are often computed indirectly from moments. For example, the following relationships are commonly used \cite{AZZOUZ199555}:
\begin{align}
  c_{4,0} &= m_{4,0} - 3\,m_{2,0}^2,\\
  c_{4,1} &= m_{4,1} - 3\,m_{2,1}\,m_{2,0},\\
  c_{4,2} &= m_{4,2} - \lvert m_{2,0}\rvert^2 - 2\,m_{2,1}^2,\\
  c_{6,0} &= m_{6,0} - 15\,m_{4,0}\,m_{2,0} + 30\,m_{3,0}^2,\\
  c_{6,3} &= m_{6,3} - 9\,c_{4,2}\,m_{2,1} - 6\,m_{2,1}^2,\\
  c_{8,0} &= m_{8,0} - 28\,m_{6,0}\,m_{2,0} - 35\,m_{4,2}^2 \notag\\
          &\quad + 420\,m_{4,0}\,m_{2,0}^2 - 630\,m_{4,0}^2.
\end{align}

Here, $m_{q,p}$ represents the mixed moment of the signal, calculated as
\begin{equation}
    m_{q,p} = \frac{1}{N} \sum_{n=1}^{N} x[n]^q \, x[n]^{p-q},
\end{equation}
with $N$ being the total number of samples. This set of formulas facilitates the efficient computation of high-order cumulants, which are subsequently used as features in our classification algorithm.

\section{Method}
\label{sec:meth}
In this section, we present a novel approach for modulation classification using higher-order statistics and prompt-based LLM generation through three key stages:

\textbf{(Stage 1)} extracting comprehensive statistical summaries (e.g., moments, variance, kurtosis); \textbf{(Stage 2)} linearizing these features into a text format \cite{wang2023meditab} as
\begin{equation} \text{linearize}(x) = {[c_{k}: x_{k}]}_{k=1}^{K}, \end{equation}
where $c_k$ is the $k^{\mathrm{th}}$ feature name and $x_k$ its value; and \textbf{(Stage 3)} constructing a structured prompt combining instructions ($I$), exemplar context ($C$), the linearized statistics, and modulation directives ($S$):
\begin{equation} d = \text{LLM}\bigl(I, C,\ \text{linearize}(x),\ S\bigr). \end{equation}
This process generates concise signal descriptions to guide the LLM in selecting the appropriate modulation type (Figure~\ref{fig:method}).

\subsection{Signal Summarization}
We compute key descriptive statistics and cumulant-related metrics to obtain a robust statistical characterization of each signal. These statistical summaries serve as the foundation for both individual signal queries and for constructing exemplar contexts, reference points with known modulation labels that help the model infer the underlying patterns associated with each modulation type. Algorithm~\ref{alg:summary} details the procedure used to convert a raw signal vector $x$ into a succinct string $d$ that encapsulates its statistical properties.

\begin{algorithm}
    \caption{Signal Summarization}\label{alg:summary}
    \KwIn{\texttt{$x$} (Vector of signal values)}
    \KwOut{\texttt{$d$} (A string summarizing cumulants and descriptive statistics)}
    
    \small 
    
    \BlankLine
    \texttt{stats} $\gets$ \{ \\
    \quad \texttt{"nobs"}: $x$.\texttt{nobs},\\
    \quad \texttt{"min"}: $x$.\texttt{min},\\
    \quad \texttt{"max"}: $x$.\texttt{max},\\
    \quad \texttt{"mean"}: $x$.\texttt{mean},\\
    \quad \texttt{"variance"}: $x$.\texttt{variance},\\
    \quad \texttt{"skewness"}: $x$.\texttt{skewness},\\
    \quad \texttt{"kurtosis"}: $x$.\texttt{kurtosis} \\
    \}\;
    \BlankLine
    
    \texttt{\# Compute k-th moment and cumulant} \\
    \For{$i \gets 0$ \KwTo $7$}{
        \texttt{stats["moment\_$i$"]} $\gets$ \texttt{moment(signal, order = $i$)}\;
    }
    \BlankLine
    
    \For{$i \gets 1$ \KwTo $4$}{
        \texttt{stats["kstat\_$i$"]} $\gets$ \texttt{kstat(signal, $i$)}\;
    }
    \BlankLine
    
    \For{$i \gets 1$ \KwTo $2$}{
        \texttt{stats["kstatvar\_$i$"]} $\gets$ \texttt{kstatvar(signal, $i$)}\;
    }
    \BlankLine
    
    $d$ $\gets$ \texttt{dict\_to\_string(stats)}\;
    \Return $d$\;
\end{algorithm}

\subsection{Example Context Construction}
To provide context for the language model, we generate statistical summaries for a set of exemplar signals, one for each modulation type, using the same procedure outlined in Algorithm~\ref{alg:summary}. These summaries, paired with their known modulation labels, are formatted into reference examples. Including these exemplars in the prompt furnishes the model with concrete illustrations of the statistical characteristics defining each modulation class, thereby enhancing classification accuracy, especially for signals affected by noise or channel impairments.

\subsection{Prompt Formulation}
The final prompt is assembled by combining three components:
\begin{itemize}
    \item \textbf{Instruction Block} ($I$): A concise set of instructions that mandates the selection of a modulation type from a predefined list (e.g., ``4ASK,'' ``4PAM,'' ``8ASK,'' etc.) and emphasizes that the output must consist of a single valid answer.
    \item \textbf{Exemplar Summaries} ($C$): The formatted examples derived from the exemplar signals, which display both the overall statistical summaries and their corresponding ground-truth modulation types.
    \item \textbf{Query Section} ($\text{linearize}(x)+S$): The statistical summary for the test signal, concatenated with a directive suffix ($S$) that reinforces the modulation classes.
\end{itemize}

This carefully designed prompt minimizes ambiguity and ensures that the language model’s output adheres strictly to the allowed modulation types and an example of a prompt is shown in Table~\ref{tab:prompt}.

\subsection{Generation and Post-Processing}
Once the prompt is constructed, it is tokenized and submitted to our fine-tuned causal language model. To encourage deterministic responses, the model is configured with constrained generation parameters (e.g., a low temperature and a restricted maximum token count). After the model generates its output, post-processing is performed to remove extraneous characters and extract the predicted modulation type by matching the response against the predefined list of valid classes. Finally, the predicted modulation type is compared with the ground truth label to compute an accuracy metric.

\begin{table}[htbp]
    \centering
    \begin{tabular}{lp{0.31\textwidth}}
    \toprule
    \multicolumn{2}{c}{\textbf{Instruction Block}} \\
    \midrule
    Prompt Instructions & 
    \begin{minipage}[t]{\linewidth}
    {You are an expert quantitative analyst in wireless communication modulation. Based on your knowledge in wireless communication modulation and the detailed signal statistics provided below, determine the modulation type.\\
    1. DO NOT PROVIDE ANY ADDITIONAL INFORMATION OR CONTEXT.\\
    2. No OTHER TEXT, NO BLABBER.}
    \end{minipage} \\
    \midrule
    \multicolumn{2}{c}{\textbf{Exemplar Summaries}} \\
    \midrule
    Example 1 & {$\{$nobs: $1024$, min: $-7.3$, max: $7.5$, mean: $-0.0$, ... $\}$, Answer 1: 4ASK} \\
    Example 2 & $\{$nobs: $1024$, min: $-6.6$, max: $7.5$, mean: $-0.1$, ... $\}$, Answer 2: 4PAM \\
    Example 3 & $\{$nobs: $1024$, min: $-19.8$, max: $15.2$, mean: $0.2$, ... $\}$, Answer 3: 8ASK \\
    \multicolumn{2}{c}{\dots \quad (Additional examples not shown)} \\
    \midrule
    \multicolumn{2}{c}{\textbf{Query Section}} \\
    \midrule
    Question Sample & { $\{$nobs: $1024$, min: $-33.502$, max: $32.946$, mean: $-0.629$, ... $\}$}\\
    \addlinespace
    Answer Choices & \texttt{4ASK, 4PAM, 8ASK, \dots} \\
    \addlinespace
    \bottomrule
    \end{tabular}
    \caption{Prompt Structure for Modulation Classification: Instruction Block, Exemplar Summaries, and Query Section.}
    \label{tab:prompt}
\end{table}

\section{Experiments and Results}
\label{sec:exp}

\begin{table*}[!htbp]
    \centering
    \caption{Modulation Classification Performance Across Models, Prompt Contexts, and Model Sizes}
    \label{tab:results}
    \begin{tabular}{lcllcc}
        \toprule
        Model & Parameters & Prompt & SNR (dB) & Accuracy (\%) & Cleaned Accuracy (\%) \\
        \midrule
        Nmformer\cite{faysal2024nmformer} & - & - &  [0.5~4.5] & 71.6 & -\\
        DenoMAE\cite{faysal2025denomae} & - & - &  [0~10] &  81.3 & -\\
        DenoMAE2.0\cite{faysal2025denomae20improvingdenoisingmasked} & - & - &  [0~10] &  \bf{82.4} & - \\
        DeepSeek-R1-Distill-Qwen  & 7B    & $I+S$    & 100 & 12.6  & 14.35 \\
        DeepSeek-R1-Distill-Qwen  & 7B    & $I+S$    & [-10~10]     & 11.8  & 13.66 \\
        DeepSeek-R1-Distill-Qwen  & 7B    & $I+C+S$  & 100 & 19.3  & 54.52 \\
        DeepSeek-R1-Distill-Qwen  & 7B    & $I+C+S$  & [-10~10]     & 5.2   & 27.81 \\
        DeepSeek-R1-Distill-Qwen & 32B   & $I+S$    & 100 & 9.1   & 10.83 \\
        DeepSeek-R1-Distill-Qwen & 32B   & $I+S$    & [-10~10]     & 7.9   & 9.58 \\
        \bf{DeepSeek-R1-Distill-Qwen} & 32B   & $I+C+S$  & 100 & \bf{58.8} & \bf{61.44} \\
        \bf{DeepSeek-R1-Distill-Qwen} & 32B   & $I+C+S$  & [-10~10]     & \bf{47.8} & \bf{53.52} \\
        \bf{OpenAI's o3-mini}            & 200B  & $I+C+S$  & [-10~10]     & \bf{69.92} & \bf{72.4} \\

    \bottomrule
    \end{tabular}
    \caption*{\emph{$I+S$} refers to the scenario where the language model receives only a concise block of instructions to select the modulation type from a predefined list without any additional exemplar context. In contrast, \emph{Example $I+C+S$} incorporates exemplar summaries derived from known modulation signals into the prompt. \emph{Cleaned Accuracy} refers to the accuracy obtained after removing responses that did not identify any modulations from the predictions.}
\end{table*}

We evaluate our context-aware automatic modulation classification framework on a synthetically generated dataset. The dataset comprises signals generated under two distinct conditions: (1) ideal, completely noiseless samples and (2) samples subjected to randomly generated noise with SNRs ranging from -10 dB to 10 dB. Since the ideal, noiseless samples contain no noise, they are significantly easier to visually distinguish and classify. In contrast, as noise levels increase, classification becomes progressively more challenging. Moreover, at lower SNRs, particularly in the negative dB range, the signals become extremely difficult to visually differentiate and accurately classify. 

Our dataset comprises signals from 10 diverse modulation schemes (4ASK, 4PAM, 8ASK, 16PAM, CPFSK, DQPSK, GFSK, GMSK, OOK, OQPSK), with SNR values systematically varied from -10dB to 10dB to simulate realistic channel conditions. We evaluate our framework based on classification accuracy across different noise conditions, comparing various prompt engineering techniques and model architectures to assess both performance and generalization capability.

\subsection{Dataset Generation}
We generated our testing dataset to include 2,000 signals featuring various modulations. For the downstream classification task, we used 20 samples as context for both noisy and noiseless data. These samples are evenly distributed among the 10 modulation classes, with SNR values ranging from -10 dB to 10 dB, where applicable. This setup allows us to assess the performance of our framework under ideal conditions as well as in the presence of channel-induced noise.

\subsection{Experimental Setup}
For each data sample, the raw complex-valued signal $x_{i,t}$ is processed to extract its I/Q components, and a comprehensive statistical summary is computed as described in Section~\ref{sec:meth}. This summary is then transformed into a natural language prompt using our linearization technique. The prompt, which includes instructions $I$, context-setting prefix $C$, the linearized statistical summary, and a directive suffix $S$, is fed into our fine-tuned language model configured with deterministic decoding parameters. The model's output, restricted to a predefined list of modulation types, is compared against the ground truth to compute classification accuracy.

\subsection{Overall Classification Performance}
{{Table~\ref{tab:results} reports classification accuracy across prompt contexts, SNR conditions, and model sizes (cleaned accuracy excludes invalid or null responses). Our results highlight the impact of context and scale on AMC performance.}}

First, incorporating exemplar context into the prompt ($I+C+S$) substantially improves performance compared to using only the instruction and suffix ($I+S$). For example, with the DeepSeek-R1-Distill-Qwen-7B model, the inclusion of exemplar context increases the cleaned accuracy from 14.35\% under noiseless conditions to 54.52\%, while under noisy conditions accuracy improves from 13.66\% to 27.81\%.

Second, the effect of model size is evident. The DeepSeek-R1-Distill-Qwen-32B model, which has significantly more parameters than the 7B variant, achieves a cleaned accuracy of 61.44\% (noiseless) and 53.52\% (noisy) with the $I+C+S$ prompt, an improvement of over 40\% relative to its $I+S$ configuration. This indicates that larger models, with their enhanced representational capacity, can better exploit the rich contextual information provided by exemplar summaries.

\begin{figure}[htbp]
  \centering
  \includegraphics[width=0.9\linewidth]{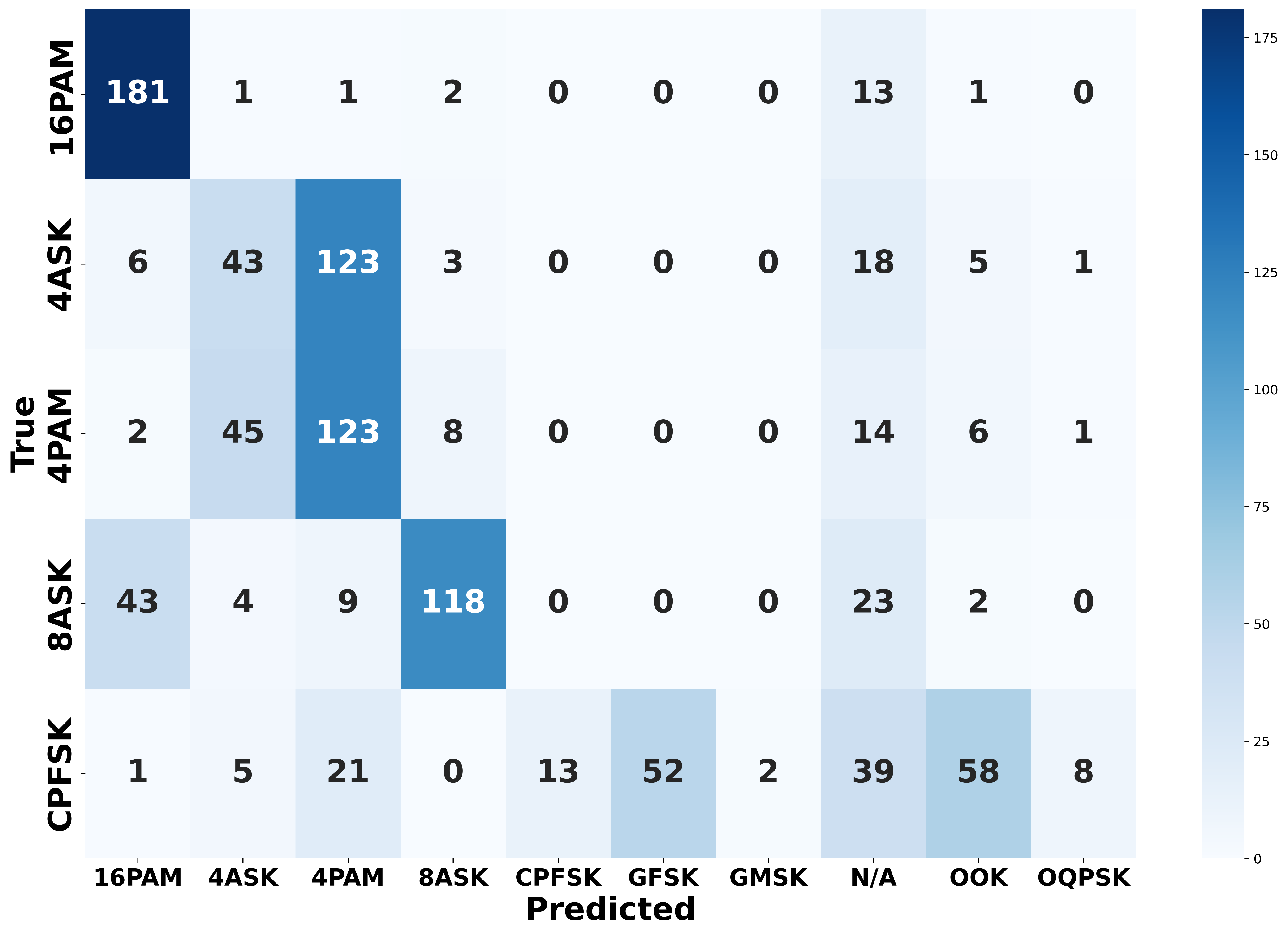}
  \caption{{Confusion matrix for DeepSeek‑R1‑Distill‑Qwen‑32B using the I+C+S prompt on noisy signals.}}
  \label{fig:confusion_deepseek32B}
\end{figure}

Figure~\ref{fig:confusion_deepseek32B} presents the confusion matrix top 5 classes for this model under noisy conditions using the $I+C+S$ prompt. While the model performs well for certain classes like 16PAM, significant confusion exists between spectrally similar modulations, notably 4ASK and 4PAM, which are frequently misclassified as each other. Other notable misclassifications include CPFSK being predicted as OOK or failing classification, and 8ASK being mistaken for 16PAM.

{{The proprietary o3‑mini model (200 B parameters) achieves the highest accuracy (69.92\% overall, 72.4\% cleaned under noise), confirming scalability benefits. Its scale, however, raises memory and latency challenges for edge or real‑time deployments, motivating future work on distillation and quantization.}}

{{Moreover, performance degrades across all models as noise increases, suggesting that explicit SNR cues could further improve robustness.}}

\section{Conclusion}
\label{sec:con}
{{

This paper presents a novel in-context approach to automatic modulation classification that takes advantage of higher-order statistics and cumulative estimation through prompt-based classification using a pre-trained large language model. By transforming quantitative signal features into natural language descriptions, our framework effectively captures the intrinsic characteristics of complex-valued signals across diverse scenarios without relying on explicit channel or SNR inputs. Experimental results on synthetically generated datasets demonstrate that incorporating exemplar context into the prompt significantly boosts classification accuracy, and that larger models—such as o3-mini can further enhance performance, particularly under noisy conditions.

Overall, our work represents a significant first step toward leveraging large language models for automatic modulation classification. We are confident that further advancements in prompt design, model architecture, and domain-specific integration will unlock new possibilities for robust and efficient signal processing in wireless communications.
}}

\FloatBarrier

\bibliographystyle{ieeetr}
\bibliography{egbib}

\begin{thebibliography}{10}

\bibitem{peng2017modulation}
S.~Peng, H.~Jiang, H.~Wang, H.~Alwageed, and Y.-D. Yao, ``Modulation classification using convolutional neural network based deep learning model,'' in {\em 2017 26th Wireless and Optical Communication Conference (WOCC)}, pp.~1--5, IEEE, 2017.

\bibitem{8963964}
T.~Huynh-The, C.-H. Hua, Q.-V. Pham, and D.-S. Kim, ``Mcnet: An efficient cnn architecture for robust automatic modulation classification,'' {\em IEEE Communications Letters}, vol.~24, no.~4, pp.~811--815, 2020.

\bibitem{10139474}
A.~Samarkandi, A.~Almarhabi, H.~Alhazmi, and Y.-D. Yao, ``Combined signal representations for modulation classification using deep learning: Ambiguity function, constellation diagram, and eye diagram,'' in {\em 2023 32nd Wireless and Optical Communications Conference (WOCC)}, pp.~1--4, 2023.

\bibitem{vaswani2017attention}
A.~Vaswani, N.~Shazeer, N.~Parmar, J.~Uszkoreit, L.~Jones, A.~N. Gomez, {\L}.~Kaiser, and I.~Polosukhin, ``Attention is all you need,'' {\em Advances in neural information processing systems}, vol.~30, 2017.

\bibitem{cai2022signal}
J.~Cai, F.~Gan, X.~Cao, and W.~Liu, ``Signal modulation classification based on the transformer network,'' {\em IEEE Transactions on Cognitive Communications and Networking}, vol.~8, no.~3, pp.~1348--1357, 2022.

\bibitem{kong2023transformer}
W.~Kong, X.~Jiao, Y.~Xu, B.~Zhang, and Q.~Yang, ``A transformer-based contrastive semi-supervised learning framework for automatic modulation recognition,'' {\em IEEE Transactions on Cognitive Communications and Networking}, 2023.

\bibitem{qu2024enhancing}
Y.~Qu, Z.~Lu, R.~Zeng, J.~Wang, and J.~Wang, ``Enhancing automatic modulation recognition through robust global feature extraction,'' {\em arXiv preprint arXiv:2401.01056}, 2024.

\bibitem{peng2018modulation}
S.~Peng, H.~Jiang, H.~Wang, H.~Alwageed, Y.~Zhou, M.~M. Sebdani, and Y.-D. Yao, ``Modulation classification based on signal constellation diagrams and deep learning,'' {\em IEEE transactions on neural networks and learning systems}, vol.~30, no.~3, pp.~718--727, 2018.

\bibitem{8454504}
F.~Meng, P.~Chen, L.~Wu, and X.~Wang, ``Automatic modulation classification: A deep learning enabled approach,'' {\em IEEE Transactions on Vehicular Technology}, vol.~67, no.~11, pp.~10760--10772, 2018.

\bibitem{hong2017automatic}
D.~Hong, Z.~Zhang, and X.~Xu, ``Automatic modulation classification using recurrent neural networks,'' in {\em 2017 3rd IEEE International Conference on Computer and Communications (ICCC)}, pp.~695--700, IEEE, 2017.

\bibitem{ramjee2019fast}
S.~Ramjee, S.~Ju, D.~Yang, X.~Liu, A.~E. Gamal, and Y.~C. Eldar, ``Fast deep learning for automatic modulation classification,'' {\em arXiv preprint arXiv:1901.05850}, 2019.

\bibitem{faysal2024nmformer}
A.~Faysal, M.~Rostami, R.~G. Roshan, H.~Wang, and N.~Muralidhar, ``Nmformer: A transformer for noisy modulation classification in wireless communication,'' in {\em 2024 33rd Wireless and Optical Communications Conference (WOCC)}, pp.~103--108, IEEE, 2024.

\bibitem{faysal2025denomae}
A.~Faysal, T.~Boushine, M.~Rostami, R.~G. Roshan, H.~Wang, N.~Muralidhar, A.~Sahoo, and Y.-D. Yao, ``Denomae: A multimodal autoencoder for denoising modulation signals,'' {\em arXiv preprint arXiv:2501.11538}, 2025.

\bibitem{faysal2025denomae20improvingdenoisingmasked}
A.~Faysal, M.~Rostami, T.~Boushine, R.~G. Roshan, H.~Wang, and N.~Muralidhar, ``Denomae2.0: Improving denoising masked autoencoders by classifying local patches,'' 2025.

\bibitem{gao2023moe}
J.~Gao, Q.~Cao, and Y.~Chen, ``Moe-amc: Enhancing automatic modulation classification performance using mixture-of-experts,'' {\em arXiv preprint arXiv:2312.02298}, 2023.

\bibitem{jafarigol2025ai}
E.~Jafarigol, B.~Alaghband, A.~Gilanpour, S.~Hosseinipoor, and M.~Mirmozafari, ``Ai/ml-based automatic modulation recognition: Recent trends and future possibilities,'' {\em arXiv preprint arXiv:2502.05315}, 2025.

\bibitem{AZZOUZ199555}
E.~Azzouz and A.~Nandi, ``Automatic identification of digital modulation types,'' {\em Signal Processing}, vol.~47, no.~1, pp.~55--69, 1995.

\bibitem{wang2023meditab}
Z.~Wang, C.~Gao, C.~Xiao, and J.~Sun, ``Meditab: scaling medical tabular data predictors via data consolidation, enrichment, and refinement,'' {\em arXiv preprint arXiv:2305.12081}, 2023.

\end{thebibliography}

\end{document}